\documentclass[11pt]{article}

\usepackage[final]{acl}
\usepackage{soul}

\usepackage{times}
\usepackage{latexsym}

\usepackage[T1]{fontenc}
\usepackage[utf8]{inputenc}

\usepackage{microtype}

\usepackage{inconsolata}

\usepackage{graphicx}
\NewDocumentCommand\gridlogo{}{\hspace*{0.73em}\raisebox{-0.54em}{\includegraphics[height=1.79em]{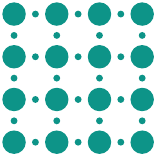}}}
\usepackage{booktabs}
\usepackage{tcolorbox}
\usepackage{enumitem}
\setlist{nosep, leftmargin=*}
\definecolor{lowcred}{RGB}{252,232,230}
\definecolor{highcred}{RGB}{226,237,250}
\definecolor{claimtopic}{RGB}{110,110,110}
\usepackage{xspace}
\definecolor{gridteal}{HTML}{0D9488}
\newcommand{\grid}{\textcolor{gridteal}{\textsc{GRID}}\xspace}
\usepackage{pdfrender}
\newcommand{\gridoutlined}{\textpdfrender{TextRenderingMode=FillStroke,LineWidth=0.35pt,LineJoinStyle=Round,StrokeColor=black,FillColor=gridteal}{\textsc{GRID}}}
\newcommand{\claimbox}[4]{%
  \begin{tcolorbox}[colback=#1, colframe=#1, boxrule=0pt, arc=1.5pt,
    left=4pt, right=4pt, top=2pt, bottom=2pt, before skip=4pt, after skip=4pt,
    sidebyside, sidebyside align=top seam, sidebyside gap=6pt,
    lefthand width=0.29\linewidth, segmentation code={},
    fontupper=\footnotesize\raggedright, fontlower=\footnotesize\raggedright]
  \mbox{\textbf{#2}}\\\textit{\textcolor{claimtopic}{#3}}
  \tcblower
  ``#4''{}
  \end{tcolorbox}\noindent\ignorespaces}

\title{\gridlogo\hspace{0.1em} \gridoutlined: Modeling Health Influencer Discourse as Argument Networks for Credibility Assessment}

\author{
Nishanth Sridhar Nakshatri, Eylon Caplan, Dan Goldwasser \\
    Purdue University \\
    {\tt \{nnakshat,ecaplan,dgoldwas\}@purdue.edu}
}

\begin{document}
\maketitle
\begin{abstract}
Health-related discussions on social media are increasingly shaped by influencers who offer diverse, and often competing perspectives on the same issue. Yet, such content is typically evaluated in isolation, without considering the broader discourse in which it appears. We introduce \grid (\textbf{G}rounded \textbf{R}easoning over \textbf{I}nterconnected \textbf{D}iscourse), a framework for evaluating influencer content through collective inference over the surrounding discourse. \grid represents each video as an argumentation graph connecting claims to its central message, or \textit{takeaway}, links related claims and takeaways across videos, grounds claims in scientific evidence, and jointly infers credibility over the resulting network. We evaluate \grid on an end-to-end health-related \textit{content validation} task using $569$ influencer videos grounded in expert assessments. Our results show that accounting for the broader discourse improves influencer content validation. Finally, we use \grid-inferred credibility to demonstrate how argumentative strategies and moral framing differ between low- and high-credibility content.
% \grid achieves $0.676$ Macro-F1, outperforming the strongest baseline by $14$ F1 points. Across various ablations, our results consistently show that accounting for the broader discourse improves automated validation of influencer content. 
% Social media influencer content is often evaluated in isolation, despite forming part of a distributed information environment in which different actors contribute complementary or competing perspectives. 
\end{abstract}

% Ablations further show that related videos and cross-video interactions contribute substantially to performance. Beyond validation, we use the resulting discourse representation to characterize how argumentative strategies and moral framing differ across content with varying levels of credibility.
% \end{abstract}

\section{Introduction}

\begin{figure}[!ht]
  \centering
  \includegraphics[scale=0.45]{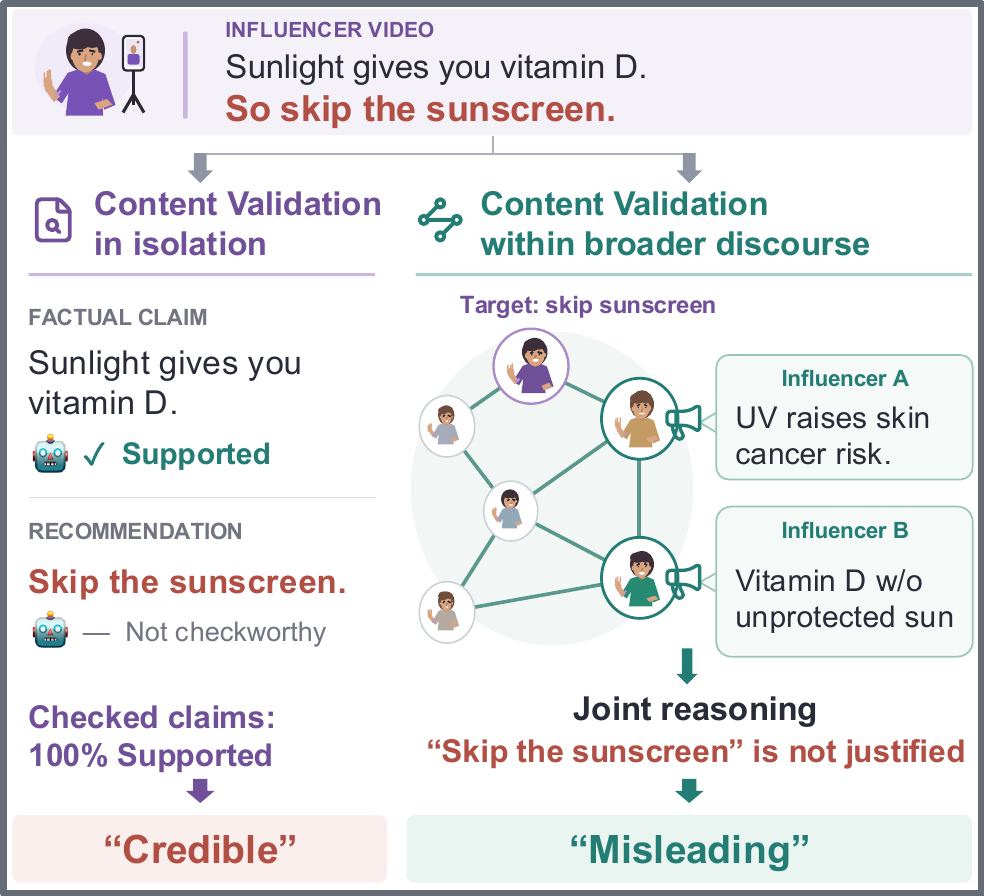}
  \caption{\small{Treating influencer content in isolation may be insufficient: a claim can be well-justified within a video yet still be incorrect. Evaluating it in isolation also ignores the broader \textit{distributed information environment}, where related influencers may provide supporting or competing perspectives. We contend that incorporating this surrounding discourse enables a more holistic evaluation of the target content.}}
  % \caption{\small{Treating influencer content in isolation fails to account for the broader \textit{distributed information environment}, where related actors may provide complementary or competing perspectives. We contend that incorporating this surrounding discourse enables a more holistic evaluation of the target content.}}
  \label{fig:figure1}
\end{figure}

Social media influencers increasingly shape how people understand and act on important issues such as health, finance, and legal matters~\cite{fabris2025self, Hayes02072024, POWELL2024116472}. Their videos often rely on using persuasive narratives~\cite{lee_superstar_2021, georgakopoulou2024influencer}, combining factual claims with lived experience, recommendations, and carefully framed justifications~\cite{MaC4717}. Such framing selectively emphasizes particular aspects of an issue to promote a product, advocate a viewpoint, or reinforce a broader philosophy. Consequently, information about an issue is often fragmented across multiple actors, each contributing different, and sometimes competing, perspectives~\cite{https://doi.org/10.1002/asi.20857}, collectively contributing to the formation of a \textit{distributed information environment}. For instance, in Figure~\ref{fig:figure1}, the target influencer video emphasizes \textit{sunlight's role in vitamin D production} to justify skipping sunscreen, while others highlight the UV exposure risks or ways to maintain vitamin D without unprotected sun, each contributing a different part of the broader picture. Viewed in isolation, the target video may appear convincing because it relies on a \textit{valid} claim while omitting other information relevant to its recommendation. Therefore, understanding and evaluating such content requires reasoning not only about how claims support a video's message, but also how that message relates to the surrounding discourse.

This view is motivated by the theory of \textit{Explanatory Coherence}~\cite{thagard1989explanatory}, which holds that hypotheses should \textit{not} be evaluated in isolation, but in relation to available evidence and competing explanations. Influencer discourse exhibits a similar structure: a video's central message, or \textit{takeaway}~\cite{alhindi-etal-2018-evidence, sundriyal-etal-2023-chaos, nakshatri-etal-2026-taigr}, is supported by claims within the video, while claims and takeaways expressed across related videos may reinforce or contradict one another. Evaluating these signals jointly therefore provides a principled way to assess how well a video's message is supported within the broader discourse.

Existing work addresses different aspects of influencer content understanding and evaluation through persuasion modeling~\cite{wang-etal-2019-persuasion}, argumentation mining~\cite{mochales2011argumentation, lawrence_argument_2019}, structured reasoning~\cite{freedman2025argumentative, zhu2025argragexplainableretrievalaugmented}, and claim verification~\cite{dmonte-etal-2026-claim}. However, these approaches typically analyze content independently, without accounting for its relationship with the surrounding discourse. In parallel, network-based approaches to news veracity detection have demonstrated the value of modeling relationships among users, posts, and claims~\cite{monti2019fake, mehta-etal-2022-tackling}. However, these approaches do not explicitly reason over the \textit{argumentative} relationships within influencer discourse---how claims and conclusions expressed across different videos support or contradict one another.

% Building on these insights, we focus specifically on the \textit{argumentative} relationships within influencer discourse---how claims and conclusions expressed across different videos support or contradict one another. Recent advances in large language models (LLMs) make it possible to explicitly model such interactions at scale. We hypothesize that jointly reasoning over these interactions enables more reliable evaluation of influencer discourse.

Building on these insights, we introduce \textbf{\grid} (\textbf{G}rounded \textbf{R}easoning over \textbf{I}nterconnected \textbf{D}iscourse), a framework for understanding and evaluating health-related influencer discourse. Given a \textit{target} video, we retrieve \textit{related} influencer videos to situate it within the surrounding discourse. We represent each video as an argumentation graph connecting its claims to its \textit{takeaway}, capturing how the influencer justifies the central message rather than treating claims independently. We then connect related claims and takeaways across videos to model the \textbf{interconnected discourse}, and \textbf{ground} its claims in external scientific evidence. Finally, we \textbf{reason} jointly over this evidence-grounded network using probabilistic inference~\cite{beltagy-etal-2014-probabilistic}, allowing evidence and argumentative relations to propagate across videos. Together, these components move beyond independent validation toward \textit{collective inference} over the broader information environment.

For evaluation, we formulate an end-to-end \textit{content validation} task that requires reasoning over influencer content to classify it as \textit{correct}, \textit{partially correct}, or \textit{incorrect}. Our dataset contains $569$ health-related influencer videos grounded in expert assessments from \texttt{ScienceFeedback}\footnote{\href{https://science.feedback.org/health-feedback/}{\nolinkurl{science.feedback.org/health-feedback}}}. \grid achieves $67.6$ Macro-F1 on this task, outperforming the strongest baseline by $+14$ F1 points.

The resulting representation also enables us to study broader patterns in how influencers construct their messages. We characterize discourse along complementary dimensions: (a) argumentative strategies~\cite{walton2008argumentation} used to support a video's \textit{takeaway}; (b) the moral values~\cite{haidt2007morality} invoked in its claims. We use these dimensions to compare communication patterns across differing levels of credibility, illustrating how our framework can support analysis of influencer discourse beyond content validation alone.

In summary, our contributions are:
\begin{itemize}
    \item We propose \grid, a framework for understanding and evaluating influencer content in the context of its surrounding discourse; jointly modeling within-video argumentation, cross-video interactions, and external scientific evidence.
    \item We show that collective inference substantially improves content validation over competing baselines (by $+14$ F1-points).
    \item We use \grid to characterize argumentative and moral framing across credibility levels, finding that \textit{authority/subversion} appears at similar overall rates but in different topics: more often in \textit{vaccine and mask} content among low-credibility videos, and in \textit{treatment and diet} content among high-credibility videos.
    % \item We use \grid to characterize argumentative and moral framing across credibility levels, finding that although low- and high-credibility content invokes the \textit{authority/subversion} foundation at similar overall rates, it does so in different topics: low-credibility videos invoke it more often in \textit{COVID-19 vaccine and mask topics}, whereas high-credibility videos invoke it more often in \textit{treatment and diet topics}.
    % similar overall rates of invoking authority mask topic-specific differences: authority is invoked more often by low-credibility videos about COVID-19 vaccines and masks, but by high-credibility videos about treatments and diet.
\end{itemize}

\section{Related Work}
Prior work relevant to understanding and evaluating influencer content has approached the problem from several complementary directions, including persuasive discourse analysis, evidence-based verification, and structured argumentation.

\begin{figure*}[!ht]
  \centering
  \includegraphics[width=\textwidth]{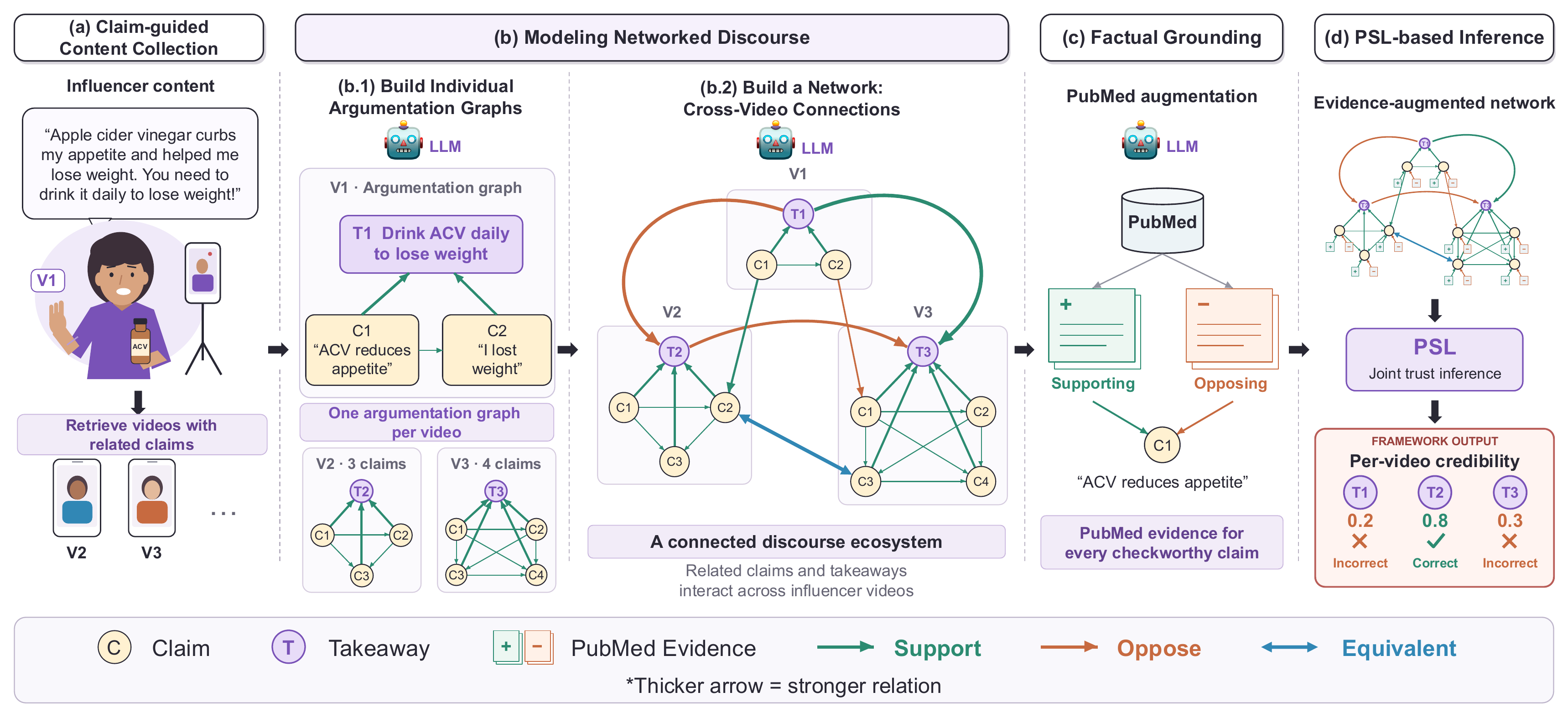}
  \caption{\small{\grid collects related influencer videos, connects their claims and takeaways in an networked argumentation graph, incorporates scientific evidence, and jointly infers per-video credibility. Note that \grid operates on influencer video transcripts.}}
  \label{fig:framework}
\end{figure*}

\textbf{Influencer and persuasive discourse.}
Prior work has examined persuasion and rhetorical strategies in online discourse~\cite{wang-etal-2019-persuasion, lee_superstar_2021, liu2024persuasive, akhynko-etal-2025-hidden, bozdag2025readsystematicsurveycomputational}, showing how messages are strategically framed to influence audiences. Related work also highlights the importance of identifying the central assertion expressed in noisy social-media content before verification~\cite{sundriyal-etal-2023-chaos}. While these studies characterize how messages are expressed or received, we focus on validating influencer content.

\textbf{Evidence-grounded and argumentative verification.}
Fact-verification systems commonly evaluate individual claims against retrieved evidence~\cite{alhindi-etal-2018-evidence, atanasova-etal-2020-generating-fact, zheng-etal-2024-evidence}, including scientific and health evidence~\cite{wadden-etal-2020-fact, kotonya-toni-2020-explainable-automated, pradeep-etal-2021-scientific, vladika-etal-2024-healthfc}. Recent LLM-based systems further combine claim decomposition, retrieval, and verification~\cite{dmonte-etal-2026-claim, li-etal-2025-loki, wang-etal-2025-openfactcheck}. Complementary approaches introduce explicit argumentative structure into verification: ArgLLM~\cite{freedman2025argumentative} uses argumentation frameworks to support interpretable verification; ArgRAG~\cite{zhu2025argragexplainableretrievalaugmented} organizes retrieved evidence through support and attack relations; MArgE~\cite{ng2025margemeshingargumentativeevidence} structures evidence from multiple LLMs into argument trees; and TAIGR~\cite{nakshatri-etal-2026-taigr} models influencer content as an argumentation graph grounded in scientific evidence. These methods, however, primarily reason within individual claims or pieces of content. We extend structured verification to the surrounding discourse by explicitly connecting claims and takeaways across related videos.

\textbf{Reasoning over broader context.}
Early work in cognitive science and psychology shows the importance of jointly reasoning over interdependent information for making informed decisions~\cite{thagard1989explanatory, THAGARD19981, FRAIDIN2004102}. Similarly, research in communication emphasizes understanding discourse through competing perspectives~\cite{https://doi.org/10.1111/j.1467-9221.2012.00873.x}. Empirical studies demonstrate the utility of broader context in argumentative discourse~\cite{durmus-etal-2019-role, zhao-etal-2021-leveraging} and misinformation detection, particularly through social interactions and propagation networks~\cite{monti2019fake, Han2020GraphNN, mehta-etal-2022-tackling}. Our work instead models the \textit{argumentative relationships} among claims and takeaways across influencer videos, grounding them in scientific evidence and jointly inferring credibility over the discourse network.

\section{\texorpdfstring{\grid}{GRID}}
Motivated by the theory of explanatory coherence, which holds that hypotheses should be assessed in relation to competing explanations and available evidence~\cite{thagard1989explanatory}, \grid validates a target video in the context of related influencer content and scientific evidence. Figure~\ref{fig:framework} summarizes its four stages: (a) retrieve related videos, (b) construct a networked argumentation graph, (c) ground its claims in scientific evidence, and (d) jointly infer video credibility through constrained optimization.

As input, \grid operates on video transcriptions generated using Whisper~\cite{radford2023robust}. The following subsections describe each stage, with implementation details in Appendix~\ref{subsec:impl-framework}.
% Our goal is to understand and evaluate health-related influencer discourse on social media. The theory of explanatory coherence in cognitive science suggests that hypotheses should be assessed in relation to competing explanations and available evidence~\cite{thagard1989explanatory}. Motivated by this view, we hypothesize that evaluating an influencer video in isolation may overlook relevant context: related videos may support its claims or offer alternative explanations, while scientific evidence provides an objective basis for evaluation. Figure~\ref{fig:framework} illustrates our four-stage framework~\textbf{\grid}: (a) collect related influencer videos, (b) represent their content as a networked argumentation graph, (c) augment the graph with scientific evidence, and (d) use constrained optimization to jointly infer the credibility of videos across the network. The following subsections describe each stage.

% Our framework operates on video transcripts and their associated captions.

\subsection{Influencer Content Collection}
\label{subsec-framework:influencer-content-collection}
Let $x_i$ denote the content of a target video by influencer $i$. To evaluate $x_i$ in context~\cite{thagard1989explanatory}, we obtain related influencer videos that may offer supporting claims or alternative explanations. Following~\cite{alhindi-etal-2018-evidence, nakshatri-etal-2026-taigr}, we guide this search by the video's \textit{takeaway}, $t_{x_i}$: the core recommendation or message that the audience is meant to retain and act upon. Focusing on the \textit{takeaway} lets us find videos about the main message itself, even when they express it through different claims. We use an LLM to identify $t_{x_i}$ and two associated keywords from $x_i$. Using the TikTok Research API, we perform a keyword-based search and filter the results for relevance to obtain the set of related videos $\mathcal{Y} = \{a_i, b_j, c_k, \ldots\}$. In the example in Figure~\ref{fig:framework}, the target takeaway is to \textit{drink apple cider vinegar (ACV) daily to lose weight}. The retrieved videos may offer contrasting perspectives, such as the claims that \textit{research has not proved that ACV causes meaningful weight loss} or that \textit{drinking ACV causes nausea}.
% TODO: Add a labeled appendix reference once implementation details and prompts are included.

\subsection{Modeling Networked Discourse}
Let $\mathcal{E} = \{x_i\} \cup \mathcal{Y}$ denote the ecosystem with the target video and its related videos. We therefore model $\mathcal{E}$ in two steps. Following prior work in claim-premise modeling and argumentation mining~\cite{mochales2011argumentation, sviridova_mining_2025, freedman2025argumentative}, we first map each video to an argumentation graph that represents the claims used to support its takeaway (see Figure~\ref{fig:framework}--b.1). We then connect related claims across these individual graphs to capture how videos support or challenge one another. The resulting networked argumentation graph (Figure~\ref{fig:framework}--b.2) represents both reasoning within videos and relations across videos, allowing us to evaluate $x_i$ in the broader discourse context.

\subsubsection{Individual Argumentation Graphs}
\label{subsubsec-framework:individual-arg-graphs}
For each video $z \in \mathcal{E}$ with takeaway $t_z$, we use an LLM to extract claims $\mathcal{C}_z = \{c_1,\ldots,c_m\}$ from its transcript and caption. For each claim, we also identify its \textit{specificity} (high or low) and \textit{source invoked} (institutional/scientific evidence, experiential evidence, social testimony, or unsupported assertion). \textit{Specificity} captures whether a claim makes concrete, bounded assertions, while \textit{source} captures the evidentiary basis invoked. These attributes characterize how a claim is presented, and provide priors for trust inference (Section~\ref{subsubsec:trust-inference}).

To capture how these claims jointly justify the video's \textit{takeaway}, we organize them into a directed argumentation graph $G_z=(V_z,A_z)$, where $V_z=\{t_z\}\cup\mathcal{C}_z$. Edges represent either direct support from a claim to the takeaway, $(c,t_z)$, or support between claims, $(c,q)$. Modeling both allows us to capture reasoning that reaches the takeaway directly or through intermediate claims.

\textbf{Graph Construction:} For each claim $c$, we first consider a candidate edge $(c,t_z)$. An LLM classifies the relation as \{\textit{direct support}, \textit{weak support}, \textit{none}\}. Using token log-probabilities~\cite{Kadavath2022LanguageM}, we score the edge as $w(c,t_z)=P(\mathrm{direct}\mid c,t_z)+0.5P(\mathrm{weak}\mid c,t_z)$ and retain it when the score exceeds a threshold. We then expand the graph recursively. Each claim connected to the takeaway becomes a new target node, and the remaining unattached claims are evaluated as potential supporters of that target using the same procedure. Expansion continues until no connected claim remains to be explored.

\subsubsection{Networked Argumentation Discourse}
\label{subsubsec:network-arg-graph}
Individual graphs capture how each influencer justifies their takeaway, but remain disconnected from arguments expressed in other videos. To represent this broader discourse, we connect them into a networked argumentation graph $G_{\mathcal{E}}=(V_{\mathcal{E}},A_{\mathcal{E}})$, where $V_{\mathcal{E}}=\bigcup_{z\in\mathcal{E}}V_z$ and $A_{\mathcal{E}}$ contains the within-video edges $\bigcup_{z\in\mathcal{E}}A_z$ plus cross-video edges encoding equivalence, support, or contradiction.

% $G_{\mathcal{E}} = (V_{\mathcal{E}}, A_{\mathcal{E}})$, where $V_{\mathcal{E}} = \bigcup_{z \in \mathcal{E}} V_z$ and $A_{\mathcal{E}}$ contains both the within-video edges $\bigcup_{z \in \mathcal{E}} A_z$ and a set of cross-video edges. The latter captures cases where videos convey equivalent messages, support one another, or express contradictory claims.

% These surrounding arguments provide the context needed to assess the target video within the broader discourse. We therefore connect the individual graphs into a networked argumentation graph 

% We model cross-video relations between both takeaways and claims. We compare all cross-video takeaway pairs to capture relations between central messages, even when they rely on different supporting claims. Because exhaustive claim comparison is costly and produces mostly irrelevant pairs, we embed claims\footnote{\texttt{all-MiniLM-L6-v2}}, cluster them using BERTopic~\cite{grootendorst2022bertopic}, and retain for each claim the top-$20$ most similar claims from other videos in its cluster.

We identify cross-video relations at two levels. First, we compare takeaways across all videos in $\mathcal{E}$, capturing relations between their central messages even when they rely on different supporting claims. Second, we connect claims across individual graphs. To avoid exhaustively comparing all cross-video claim pairs, we embed claims with \texttt{all-MiniLM-L6-v2}~\cite{reimers-gurevych-2019-sentence} and cluster semantically related claims using BERTopic~\cite{grootendorst2022bertopic}. For each claim, we consider the top-$20$ most similar claims from other videos from the same cluster.
% For small claim sets and clustering outliers, we directly select the most similar cross-video claims. This candidate-selection step narrows the comparison set while retaining claims likely to participate in meaningful relations.

For each candidate pair $(u,v)$, an LLM classifies the directed relation from $u$ to $v$ as \textit{equivalent}, \textit{supports}, \textit{contradicts}, or \textit{none}. Similar to individual graph construction, we derive class probabilities from the model's token log-probabilities and assign the edge score: $w(u,v) = P(\mathrm{supports}\mid u,v) + P(\mathrm{equivalent}\mid u,v)
- P(\mathrm{contradicts}\mid u,v).$
% Equivalent claims express the same proposition; support indicates that $u$ provides a direct reason to accept $v$; contradiction indicates that $u$ rejects or conflicts with $v$; and none indicates that the pair has no meaningful argumentative relation. As with the individual graphs, we derive class probabilities from the model's token log-probabilities and assign the edge score

For every relation classified as other than \textit{none}, we add a cross-video edge with its relation type and score. Positive scores encode support or equivalence, whereas negative scores encode contradiction. The resulting graph combines the reasoning expressed within individual videos with relations across their takeaways and claims, providing the connected discourse structure used for subsequent evidence augmentation and joint trust inference.

\subsubsection{Factual Grounding}
\label{subsubsec-framework:factual-grounding}
The networked graph $G_\mathcal{E}$ (Section~\ref{subsubsec:network-arg-graph}) represents influencer discourse but does not establish factual validity. We therefore augment it with scientific evidence from PubMed, following the evidence-retrieval pipeline in~\cite{li-etal-2025-loki}.

\textbf{Checkworthy nodes.}
Since not every claim is verifiable against biomedical literature, we use an LLM to identify \textit{PubMed-checkworthy} claims. Only these undergo evidence retrieval; all other claims remain in $G_\mathcal{E}$.

\textbf{Query expansion and retrieval.}
For each checkworthy claim, an LLM generates up to $5$ supporting and $5$ opposing queries to retrieve evidence both \textit{for} and \textit{against} it. Using \texttt{ColBERTv2}~\cite{santhanam-etal-2022-colbertv2}, we retrieve the top-$20$ PubMed documents per query, aggregate them across queries, and retain the top-$20$ documents in each supporting and opposing pool. An LLM then filters these documents for relevance.

\textbf{Evidence classification and graph augmentation.}
An LLM classifies each candidate as \{\textit{strongly supporting} (SS), \textit{weakly supporting} (WS), \textit{weakly opposing} (WO), \textit{strongly opposing} (SO), \textit{neutral}, or \textit{irrelevant}\}. We retain up to three articles per pool, prioritizing stronger evidence (SS or SO) and excluding neutral or irrelevant articles. Each retained article $e$ is added as a node linked to claim $c$ by an evidence edge $(e,c)$ with score
$w(e,c)=P(\mathrm{SS}\mid e,c)+0.5P(\mathrm{WS}\mid e,c)-0.5P(\mathrm{WO}\mid e,c)-P(\mathrm{SO}\mid e,c).$
The score lies in $[-1,1]$, with positive values indicating support and negative indicating opposition. The evidence-augmented graph is then used to jointly infer trust and estimate target video's credibility.

\subsubsection{Trust Inference}
\label{subsubsec:trust-inference}
Given the evidence-augmented networked argumentation graph, we jointly infer trust over its claims and takeaways using probabilistic soft logic (PSL)~\cite{bach:uai13, 3122009.3176853}. PSL assigns soft truth values in $[0,1]$ and captures relational dependencies through first-order logical rules~\cite{beltagy-etal-2014-probabilistic}. Let $T(v)\in[0,1]$ denote the trust of node $v$. We treat retrieved PubMed evidence as trusted, fixing $T(e)=1$, while jointly inferring trust for claims and takeaways.

Each claim $c$ is also assigned a prior $\pi(c)=s(c)b(c)$ based on its extracted \textit{specificity} and \textit{source} attributes.\footnote{\footnotesize{We use $s(c) \in \{1,0.5\}$ for high/low specificity and $b(c)\in\{1,0.5,0.25,0.1\}$ for institutional/scientific, experiential, social testimony, and unsupported sources, respectively.}} These priors capture how a claim is presented rather than its factual validity; takeaways receive no metadata-based prior. Let $S(u,v)$, $A(u,v)$, and $E(u,v)$ denote observed support, attack, and equivalence relations, with values given by their edge score magnitudes. We define:
\[
\begin{array}{rcl}
R1: T(u)\land S(u,v) &\rightarrow& T(v),\\
R2: T(u)\land A(u,v) &\rightarrow& \neg T(v),\\
R3: E(u,v)\land T(u) &\rightarrow& T(v),\\
R4: E(u,v)\land T(v) &\rightarrow& T(u),\\
\multicolumn{3}{c}{R5: \Pi(c)\rightarrow T(c),\quad R6: \neg\Pi(c)\rightarrow\neg T(c),}
\end{array}
\]
where $\Pi(c)$ is observed with value $\pi(c)$. $R1$--$R2$ propagate support and attack from trusted nodes, $R3$--$R4$ encourage equivalent nodes to receive consistent trust, and $R5$--$R6$ softly anchor claims to their metadata priors.

Under PSL's soft-logic relaxation, each grounded rule $r$ induces a hinge-loss potential $\phi_r(\mathbf{T})=\max\{0,\ell_r(\mathbf{T})\}^{p_r}$,
where $\ell_r(\mathbf{T})$ is its linear rule violation and $p_r\in\{1,2\}$. MAP inference then finds the joint trust minimizing total weighted rule violation~\cite{3122009.3176853}:
\[
\mathbf{T}^{*}
=
\mathop{\mathrm{arg\,min}}_{\mathbf{T}\in[0,1]^n}
\sum_{r\in\mathcal{R}} \lambda_r \phi_r(\mathbf{T}).
\]
We use unit rule weights and linear hinge loss ($\lambda_r$=$1$, $p_r$=$1$). The resulting $\mathbf{T}^{*}$ jointly resolves scientific evidence, discourse relations, and claim priors over the network. We use $T^{*}(t_z)$ as the credibility score of video $z$ and map it to \textit{correct}/\textit{partially correct}/\textit{incorrect} with fixed thresholds.

\section{Evaluation Task}
We evaluate \grid through a \textit{content validation} task: given an influencer content, classify it as \{\textit{correct}, \textit{partially correct}, \textit{incorrect}\}. We choose this task because successful prediction requires both understanding how the video's claims support its message and evaluating those claims against evidence, providing an end-to-end evaluation.
% Having described how \textbf{\grid} understands and evaluates influencer discourse, we now test its efficacy through a \textit{content validation} task. Given the transcript of an influencer video, the goal is to classify its content as {\textit{correct}, \textit{partially correct}, \textit{incorrect}}. This task requires both understanding the claims and reasoning used to support the video's message, and evaluating them against evidence. It therefore provides a holistic measure of the ability to characterize and assess influencer content.

% The task requires identifying and interpreting the claims and reasoning expressed in the target video. Based on this understanding, the video content is classified as {\textit{correct}, \textit{partially correct}, \textit{incorrect}}. We use this end-to-end task because successful prediction depends on both: understanding how the influencer supports their message and evaluating those arguments against relevant evidence; providing a holistic measure to validate our approach.
% Content-validation performance therefore provides a holistic measure of the effectiveness of our approach.

% \footnote{\href{https://science.feedback.org/health-feedback/}{\nolinkurl{science.feedback.org/health-feedback}}}
\subsection{Annotated Dataset}
\textit{Content validation} task requires gold-standard judgments of factual validity. We therefore construct an evaluation set from TikTok videos whose central message (\textit{takeaway}) aligns with expert-annotated claims from \texttt{ScienceFeedback}. Medical experts on \texttt{ScienceFeedback} assign health-related claims one of $18$ fine-grained veracity labels (Appendix~\ref{subsec:impl-annotated-dataset}). As these are sparsely represented, we map them to our three task labels: \{\textit{correct}, \textit{partially correct}, \textit{incorrect}\}.

For each expert-annotated claim, we extract up to two representative keywords and retrieve candidate videos through the TikTok Research API. Because topical similarity alone does not guarantee that the expert judgment applies to a video's \textit{takeaway}, we first filter candidates for relevance and then use an LLM to verify alignment between the video's takeaway and the corresponding \texttt{ScienceFeedback} claim. Aligned videos inherit the expert veracity label. We validate this procedure through \textbf{human evaluation} on $90$ randomly sampled videos: extracted takeaways are judged correct for $91\%$ of videos, with $90\%$ annotator agreement, while $84\%$ align with the corresponding \texttt{ScienceFeedback} claim, with $80\%$ agreement (Appendix~\ref{sec:human-evaluation}). We further restrict retrieval to creators with at least $1{,}000$ followers to focus on influencers with an established audience.

The aligned videos serve as labeled target instances, while unlabeled related videos are collected using the procedure in \S~\ref{subsec-framework:influencer-content-collection} to provide surrounding discourse. The final dataset contains \textbf{1245} videos from \textbf{1050} unique influencers: \textbf{569} labeled target videos and \textbf{676} unlabeled related videos.

\section{Experiments}
We evaluate \grid on the \textit{content validation} task against several baselines and conduct ablations to validate the contribution of its components. Unless otherwise stated, all methods use \texttt{gpt-5.4-mini}~\cite{openai2026gpt54} for takeaway and claim extraction, and \texttt{Qwen3.5-9B}~\cite{yang2025qwen3} for all remaining LLM-based components. We report results from a single model run.

\subsection{Baselines}
\label{sec:baselines}
We compare \grid against four baselines:

(a) \textbf{Direct Prompting:} We prompt \texttt{Qwen3.5-9B} with the target transcript to directly predict \{\textit{correct}, \textit{partially correct}, \textit{incorrect}\}, relying only on the model's parametric knowledge. 

(b) \textbf{LOKI:} We adapt LOKI~\cite{li-etal-2025-loki} to our task. LOKI extracts a flat set of checkworthy claims, retrieves evidence, verifies each claim, and aggregates claim-level scores into a video-level prediction. 

(c) \textbf{Per-video Pragmatic Inference:}
Following~\citet{nakshatri-etal-2026-taigr}, we apply our structured reasoning pipeline to the target video independently: construct its argumentation graph, ground claims with PubMed evidence, and infer takeaway trust. This captures within-video argumentative structure but excludes cross-video relations. 

(d) \textbf{Per-video Score Aggregation:}
We independently apply the same per-video inference to the target and related videos, then average their takeaway trust scores; serving as a continuous version of \textit{majority voting}. This tests whether aggregating predictions across related videos is sufficient without explicitly modeling cross-video interactions.

For the latter two baselines, trust scores are mapped to classes using tuned thresholds. Implementation details are provided in Appendix~\ref{subsec:impl-baselines}.
% We compare our framework against the following baselines.

% \textbf{Direct Prompting.}
% We directly prompt \texttt{Qwen3.5-9B} with the target video's transcript and ask it to classify the content as \textit{correct}, \textit{partially correct}, or \textit{incorrect}. This baseline measures performance using only the LLM's parametric knowledge, without explicit discourse modeling or external evidence retrieval. Implementation details are provided in Appendix~\ref{subsec:impl-baselines}.

% \textbf{LOKI.}
% We adapt LOKI~\cite{li-etal-2025-loki} to the content validation task. LOKI decomposes the video into a flat set of claims, identifies checkworthy claims, retrieves PubMed evidence through query expansion, verifies individual claims against the retrieved evidence, and aggregates claim-level scores to obtain a video-level prediction. Implementation details are provided in Appendix~\ref{subsec:impl-baselines}.

% Main results table: compact one-column version (scores x100).
% Alternatives kept below inside \iffalse ... \fi: full-width x100 version and original 0-1 version.
\begin{table}[t!]
\centering
\small
\setlength{\tabcolsep}{4pt}
\begin{tabular}{@{}lc@{\hspace{\tabcolsep}\textcolor{black!25}{\vrule width 0.4pt}\hspace{\tabcolsep}}ccc@{}}
\toprule
& & \multicolumn{3}{c}{\textbf{Class-wise F1}} \\
\cmidrule(l){3-5}
\textbf{Method} & \textbf{Macro F1} & \textit{Inc.} & \textit{Part.} & \textit{Corr.} \\
\midrule
Direct prompting & 45.2 & 86.9 & 12.3 & 36.4 \\
Per-video pragmatic & 46.0 & 82.1 & 18.4 & 37.6 \\
Per-video aggregation & 53.1 & 89.8 & \phantom{0}6.2 & \textbf{63.5} \\
LOKI~\cite{li-etal-2025-loki} & 30.8 & 52.8 & 14.5 & 25.0 \\ % TODO: Add LOKI-style results.
\midrule
\textbf{\grid (ours)} & \textbf{67.6} & \textbf{92.1} & \textbf{50.8} & 60.0 \\
\bottomrule
\end{tabular}
\caption{\small{Main results for content validation: \textbf{\grid} outperforms all baselines in Macro-F1, showing the benefit of the context of broader influencer discourse. Inc./Part./Corr.: \textit{incorrect}/\textit{partially correct}/\textit{correct}.}}
\label{tab:main-results}
\end{table}

\begin{figure}[b]
  \centering
  \includegraphics[width=\columnwidth]{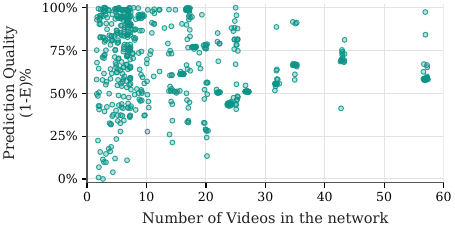}
  \caption{\small{\textbf{Larger networks protect \grid from very poor predictions.} In networks of up to $20$ videos, prediction quality spans nearly the full range from $0$ to $100\%$; in larger networks, every prediction stays above $40\%$. Points show individual target videos, plotted by number of videos in their network.}}
  \label{fig:predqual_vs_networksize_sensitivity}
\end{figure}

\subsection{Results}
Extended results are reported in Appendix~\ref{subsec:extended-ablation}.

\textbf{Main Results.} Table~\ref{tab:main-results} presents results on the \textit{content validation} task. \grid outperforms the strongest baseline by $14$ F1-macro points. The claim-centric LOKI baseline performs substantially worse, partly due to limited coverage~\cite{nakshatri-etal-2026-taigr}: it identifies checkworthy claims for only $\approx75\%$ of target videos, restricting the evidence available for validation. Modeling the target video as an argumentation graph with \textit{per-video pragmatic inference} improves F1-macro, showing the benefit of reasoning over its internal argumentative structure. Incorporating related videos through score aggregation yields a further $7$-point gain, indicating that the surrounding discourse provides useful context. Our full method performs best by explicitly modeling interactions across videos, supporting our premise that influencer content should be evaluated jointly within the broader discourse rather than in isolation.

 % Note that coverage from LOKI is $\approx 75\%$ (vs.\ $100\%$ for other methods); Lack of coverage from LOKI is due to its inability to identify checkworthy nodes or retrievable evidence.

 \begin{figure}[t]
  \centering
  \includegraphics{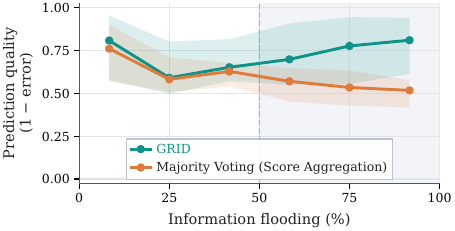}
  \caption{\small{\textbf{Does \grid conform to the majority view when flooded with information?} While majority voting's prediction degrades with information flooding, \grid maintains and even improves in content validation.}}
  \label{fig:info_flood_predqual_vs_networksize}
\end{figure}

\textbf{Impact of Network Size.} We examine how the number of related videos affects prediction quality. Figure~\ref{fig:predqual_vs_networksize_sensitivity} plots prediction quality, defined as $1-E$ ($E$ is the absolute distance between the predicted trust score and the numeric gold score), against \textit{network size}, measured as the number of videos in the network including the target video and its related videos. For smaller network size, prediction quality varies widely, spanning nearly the full range from $0$ to $1$. As network size increases, \grid is protected from bad predictions, suggesting that related videos provide useful context for inference. We further test this by randomly removing $25\%$, $50\%$, and $75\%$ of related videos; performance consistently decreases as more videos are removed (Table~\ref{tab:related-video-removal}).

% We want to test the overall impact of having related videos. Figure~\ref{fig:predqual_vs_networksize_sensitivity} shows the plot, with prediction quality ($1-E$, where $E$ is the absolute distance between the predicted trust score and the numeric gold score) on Y-axis and network size (calculated as the total number of related videos along with the target video) on the X-axis. With smaller network size, we observe a higher prediction quality range, ranging between 0 to 100\%, whereas this range reduces with increasing network size; validating that predictions from our method improve by considering more related influencer videos. To further validate this empirically, we randomly remove 25\%, 50\%, 75\% of the related videos and observe consistent performance drop with each removal, as outlined in Table~\ref{tab:related-video-removal}. Additional details are found in Appendix~ref{}.

% \textbf{Composition of related videos.} 
\textbf{Varying Network Composition.}
We vary the composition of the related-video network by controlling the proportion of videos that oppose the target, which we refer to as \textit{information flooding}. At $0\%$ flooding, all related videos express a perspective consistent with the target; as flooding increases, opposing perspectives increasingly dominate the network. This allows us to test whether our networked inference merely tracks the dominant signal in the neighborhood, as simple score aggregation over neighbors would, or instead reasons over the structure of cross-video interactions.

Figure~\ref{fig:info_flood_predqual_vs_networksize} shows the median prediction quality, with the $25$--$75\%$ IQR band, across flooding levels. We observe that our method and \textit{Per-video Score Aggregation} increasingly diverge once flooding exceeds $50\%$, when opposing perspectives become the majority. While score aggregation degrades under this shift, our method remains comparatively robust, indicating that its predictions are not determined by simply aggregating per-video scores.
% First, the median prediction quality is relatively stable across flooding levels for the smallest networks ($2-10$ related videos), although individual predictions remain highly dispersed, as reflected by the wide IQR band. For larger networks, this band narrows, indicating concentrated model predictions, but the median prediction quality changes more noticeably at some flooding levels.

% Particularly, for larger networks, prediction quality increases at high flooding levels: beyond $\approx 70\%$ flooding for networks of size $20-30$, and beyond $\approx 80\%$ for networks larger than $30$. Importantly, these regions contain only targets labeled \textit{incorrect} in our dataset; no partially-correct or correct targets occur at these network sizes. Thus, this trend should not be interpreted as a general benefit of increased flooding. Rather, in these cases, increased flooding exposes an incorrect target to more opposing (often correct) surrounding discourse, providing useful contrastive evidence that helps our collective inference framework identify the target as incorrect more accurately.

\textbf{Impact of different interactions.} To evaluate the contribution of different edge types, we remove (a) takeaway--takeaway, (b) cross-video claim--claim, and (c) PubMed--claim connections, while retaining all other edges. As shown in Table~\ref{tab:interaction-ablation}, removing cross-video claim--claim connections reduces F1-macro by $\approx 7$ points, while removing takeaway--takeaway connections causes a larger $16$-point drop. These results show that explicit cross-video interactions, particularly between takeaways, are critical: \grid's gains go beyond simply conforming to the majority.
% , as in the \textit{Per-video Score Aggregation} baseline.

% Interaction ablation table: compact one-column version (scores x100).
% Alternatives kept below inside \iffalse ... \fi: the full-width table* and the original one-column table.
\begin{table}[t!]
\centering
\small
\setlength{\tabcolsep}{4pt}
\begin{tabular}{@{}lc@{\hspace{\tabcolsep}\textcolor{black!25}{\vrule width 0.4pt}\hspace{\tabcolsep}}ccc@{}}
\toprule
& & \multicolumn{3}{c}{\textbf{Class-wise F1}} \\
\cmidrule(l){3-5}
\textbf{Removed connection} & \textbf{Macro F1} & \textit{Inc.} & \textit{Part.} & \textit{Corr.} \\
\midrule
Takeaway--takeaway & 51.4 & 84.5 & 27.5 & 42.2 \\
Cross-video claim--claim & 60.9 & 90.3 & 34.9 & 57.4 \\
PubMed--claim & 37.8 & 85.1 & \phantom{0}4.7 & 23.5 \\
\bottomrule
\end{tabular}
\caption{\small{\textbf{Every connection type contributes, and scientific evidence contributes most.} F1 (\%) after removing each connection type; full \grid reaches $67.6$ Macro F1 (Table~\ref{tab:main-results}). Without PubMed--claim connections, Macro F1 drops by $29.8$ points and \textit{partially correct} F1 nearly vanishes. Inc./Part./Corr.: \textit{incorrect}/\textit{partially correct}/\textit{correct}.}}
\label{tab:interaction-ablation}
\end{table}

These results show that: (i) related videos provide useful context for evaluating the target, and (ii) explicitly modeling cross-video interactions is more effective than simple score aggregation, even when opposing perspectives dominate the network.
% (iii) scientific grounding remains critical for reliable validation.
% \textbf{Human Evaluation.}

\section{Analysis of Argumentative \& Moral Framing by Credibility}
\label{sec:discourse-characterization}

\begin{figure}[t]
  \centering
  \includegraphics{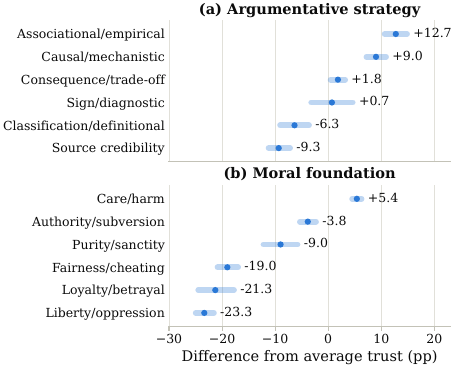}
  \caption{\small{{\textbf{Which framings are most credible?} Claims citing empirical data or causal mechanisms, and claims framed around care/harm, receive above-average trust, while claims framed around fairness, loyalty, or liberty receive the least. Points show mean claim trust relative to the mean over all claims with $95\%$ bootstrap intervals.}}}
  \label{fig:trust-by-discourse-category}
\end{figure}

% \nish{need to rewrite ...}
The trust scores inferred by \grid also enable analysis of how influencers argue and frame their claims, including unlabeled videos without expert judgments. We use an LLM to label each of the $13{,}717$ extracted claims with at most one argumentative strategy and any number of moral foundations (definitions in Table~\ref{tab:discourse-label-definitions}; human validation in Appendix~\ref{sec:human-evaluation}), and relate these labels to claim trust $T^{*}(c)$ and the credibility $T^{*}(t_z)$ of the corresponding video.

\begin{table*}[t]
\centering
\small
\setlength{\tabcolsep}{5pt}
\renewcommand{\arraystretch}{1.05}
\begin{tabular}{@{}p{0.19\textwidth}p{0.27\textwidth}p{0.19\textwidth}p{0.27\textwidth}@{}}
\toprule
\multicolumn{2}{c}{\textbf{Argumentative strategy}} & \multicolumn{2}{c}{\textbf{Moral foundation}} \\
\cmidrule(lr){1-2}\cmidrule(lr){3-4}
\textbf{Label} & \textbf{Definition} & \textbf{Label} & \textbf{Definition} \\
\midrule
Causal/mechanistic & Explains a causal process connecting the claim to the takeaway. & Care/harm & Emphasizes preventing suffering, injury, or illness. \\
Associational/empirical & Uses an observed association, measurement, or empirical result. & Fairness/cheating & Invokes equitable treatment, honesty, or accountability. \\
Consequence/trade-off & Reasons from expected outcomes or balances costs and benefits. & Loyalty/betrayal & Invokes obligations to, or betrayal of, a group. \\
Classification/definitional & Defines or categorizes an entity, event, or intervention. & Authority/subversion & Defers to or challenges institutions, expertise, or hierarchy. \\
Sign/diagnostic & Treats an observation as a sign of an underlying condition. & Purity/sanctity & Invokes contamination, naturalness, or bodily integrity. \\
Source credibility & Supports or challenges a claim through a source's expertise, trustworthiness, or conflicts. & Liberty/oppression & Invokes autonomy, coercion, or restrictions on choice. \\
\bottomrule
\end{tabular}
% \nish{read through}
\caption{Argumentative-strategy and moral-foundation labels used in the discourse analysis.}
\label{tab:discourse-label-definitions}
\end{table*}

% The trust scores inferred by \textbf{\grid} also let us study how influencers argue and frame their claims, including in the unlabeled videos for which no expert judgment exists. We use an LLM to label each of the $13{,}717$ claims extracted from our dataset with at most one argumentative strategy and any number of moral foundations (definitions in Appendix Table~\ref{tab:discourse-label-definitions}, human validation of labels in Appendix~\ref{sec:human-evaluation}), and relate these labels to the inferred claim trust $T^{*}(c)$ and to the credibility score $T^{*}(t_z)$ of the video containing the claim.

\textbf{Moral foundations separate claim trust more than argumentative strategies.}
Figure~\ref{fig:trust-by-discourse-category} reports the mean trust of claims in each category relative to the mean over all claims. Among argumentative strategies, associational/empirical and causal/mechanistic claims receive the highest trust ($+12.7$ and $+9.0$ points), and source credibility claims the lowest ($-9.3$). \textbf{Differences across moral foundations are larger.} Care/harm is the only foundation above the mean ($+5.4$), while claims invoking fairness/cheating, loyalty/betrayal, or liberty/oppression fall $19$--$23$ points below it. These claims typically concern rights, mandates, or unfair treatment, which the biomedical literature rarely addresses, so they are seldom linked to supporting evidence. When the two dimensions are combined, \textbf{the moral foundation matters more than the strategy}: causal/mechanistic claims are $11.5$ points above the mean when they invoke care/harm, but $18.2$ points below it when they invoke fairness/cheating (Appendix Figure~\ref{fig:strategy-foundation-trust}). Health claims that invoke fairness, loyalty, or liberty therefore tend to lack evidential support regardless of how they are argued.

\begin{figure}[t]
  \centering
  \includegraphics{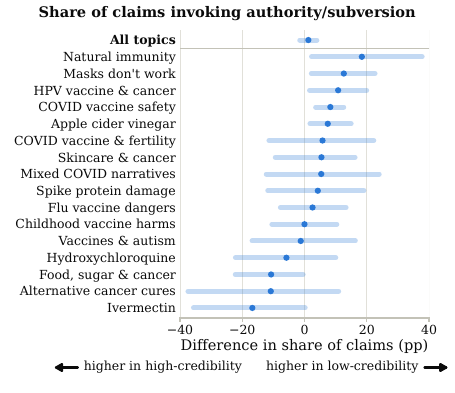}
  \caption{\small{{\textbf{Do low-credibility videos invoke authority more often?} On average, no (all topics $\approx 0$). But in \textit{COVID-19 vaccine} and \textit{mask} topics they invoke authority more, whereas in \textit{treatment} and \textit{diet}, high-credibility videos do.}}}
  \label{fig:authority-by-topic}
\end{figure}

\textbf{Authority is invoked by both low- and high-credibility videos.}
The \textit{authority/subversion} foundation includes both deferring to and challenging institutions. To examine how its use relates to credibility, we manually group the $145$ expert-reviewed claims in our dataset into health topics and, within each topic, compare the share of claims invoking authority in low-credibility ($T^{*}(t_z)\leq 0.5$) and high-credibility videos (Figure~\ref{fig:authority-by-topic}). Across all topics, the two groups invoke authority at similar rates ($23\%$ vs.\ $22\%$ of claims), but this hides a split by topic: \textbf{in some topics, appeals to authority are more common among credible influencers, while in others they are more common among influencers who should not be trusted.} In topics about COVID-19 vaccines and masks, authority is more common in low-credibility videos (by $8$--$18$ points), where it is typically used to challenge institutions:
\claimbox{lowcred}{Low-credibility}{Natural immunity}{Experts and the media hid evidence that natural immunity protects against severe COVID-19.}
In topics about treatments and diet, such as ivermectin, it is more common in high-credibility videos (by $11$--$17$ points), where it is typically used to cite institutions:
\claimbox{highcred}{High-credibility}{Ivermectin}{A 2021 Cochrane review of 11 studies found no benefit of ivermectin for COVID-19.}
The same moral foundation can thus serve opposite purposes in different topics, so its presence does not indicate whether content should be believed.
\section{Conclusion}
We introduced a structured framework, \grid, which validates influencer content through collective inference over interconnected discourse and scientific evidence. \grid outperforms competing baselines on content validation and enables analysis of how discourse patterns vary with credibility.
\section*{Ethical Considerations}
\paragraph{Datasets.}
Our work analyzes publicly available health-related influencer content retrieved through the TikTok Research API and expert assessments published by ScienceFeedback.

\paragraph{Privacy and identifiability.}
Although the analyzed content is public, health communication can involve sensitive topics and identifiable individuals. We therefore do not infer private attributes. To reduce privacy risks associated with multimedia data, our model operates only on textual transcripts and do not use or release raw video, audio, images, or biometric signals such as faces or voices. We also avoid reporting identifying information about individual creators and present our analyses in aggregate.

\paragraph{Content-level evaluation.}
Our framework evaluates the credibility of video content, not the overall trustworthiness, expertise, or intent of the creator. A prediction that a video is \textit{incorrect} or \textit{partially correct} should therefore \textbf{not} be interpreted as a characterization of the individual who produced it. This distinction is particularly important because creators may discuss different topics with varying levels of accuracy.

\paragraph{Health information and scientific evidence.}
Our framework relies on retrieved biomedical literature and expert-derived labels to evaluate health-related claims. Scientific evidence may be incomplete, contested, or change over time, and errors may arise from claim extraction, retrieval, relevance filtering, or stance classification. Moreover, not all claims---such as personal experiences or subjective recommendations---can be meaningfully evaluated using biomedical literature. We therefore restrict evidence retrieval to PubMed-checkworthy claims and do not treat the absence of retrieved evidence as evidence that a claim is false.

\paragraph{Potential misuse and deployment.}
Network-level credibility inference could be misused for automated moderation, surveillance, or ranking individuals based on model-generated credibility estimates. Our framework is intended for research on discourse understanding and content validation. It is not meant for medical decision-making. Any real-world deployment should account for model uncertainty, evolving evidence, and the potential consequences of erroneous predictions, and should incorporate appropriate human oversight.
\section*{Limitations}

\textbf{Scope and representativeness.}
Our study focuses on English-language health videos from TikTok creators with at least $1{,}000$ followers. The dataset is modest ($1{,}245$ videos, of which $569$ are labeled), and its coverage is shaped by: (a) the claims available from \texttt{ScienceFeedback}, (b) the selected search terms, and (c) the TikTok Research API. The findings may therefore not generalize to smaller creators, other platforms, languages, or domains. The framework itself can extend to settings with suitable external evidence, such as legal or financial content, but this requires domain-specific sources, label definitions, and evaluation rather than direct transfer of the present results.

\textbf{Evaluation labels.}
Our labels are transferred from expert-reviewed claims to videos whose extracted takeaways are semantically aligned; they are not direct expert assessments of the complete videos. Human evaluation finds $84\%$ takeaway--expert alignment, leaving room for transfer errors. Moreover, $46$ of the $57$ \textit{correct} examples are proxy instances aligned with curated \texttt{MayoClinic} takeaways, and the final evaluation set is imbalanced ($478$ \textit{incorrect}, $34$ \textit{partially correct}, and $57$ \textit{correct}). Mapping fine-grained source verdicts into three classes also removes distinctions among different types of error. Larger datasets with direct, multi-expert video annotations and stronger coverage of the minority classes would support more conclusive evaluation.

\textbf{Content and network coverage.}
\grid operates on transcripts and captions, so it does not model visual demonstrations, or other multimodal signals that may alter a video's meaning. Its discourse network is obtained through keyword retrieval and relevance filtering and should be understood as a sampled view of related content, rather than the complete information environment encountered by users. We also omit social and propagation context such as follower relations, comments, and engagement; moreover, the largest network in our experiments contains $57$ videos. Future work could combine multimodal content with social context for substantially larger networks.

\textbf{Evidence and modeling assumptions.}
Factual grounding is restricted to English PubMed title--abstract records published from 2014 through 2025. This provides a consistent biomedical source but may miss older work, clinical guidance, and any evidence outside PubMed. We also fix every retained PubMed node at maximum trust, which does not distinguish study design, methodological quality, retractions, or the strength of scientific consensus. Incorporating multiple curated evidence sources and estimating \textit{source-} and \textit{study-}level reliability are important extensions.

Several stages---takeaway and claim extraction, relation identification, evidence filtering, and discourse labeling---depend on LLM outputs, so errors can propagate through the graph. Human validation is encouraging but limited to $90$ videos and $270$ claims; argumentative strategies and moral foundations also have lower raw agreement ($70\%$ and $71\%$), reflecting their subjectivity. Our reported results use unit rule weights, fixed decision thresholds, two particular LLMs, and a single model run. Future work should test robustness across models, prompts, random seeds, inference settings, and independently annotated evaluation sets.

\textbf{Interpretation and use.}
The argumentative- and moral-framing results are descriptive associations with trust inferred by \grid; they do not show that a framing choice causes credibility, nor that low inferred trust establishes falsehood. Topic composition, evidence availability, and modeling choices can all affect these patterns. Finally, performance remains imperfect, particularly for the minority classes, so the framework should support human review rather than serve as an automated tool of influencer content credibility.

\section{Statement of AI Assistant Usage}
\label{sec:ai-assistant}
The development of the codebase was partially assisted by an AI system, and all AI-generated code was reviewed and verified by the authors. An AI assistant was also used to help refine the manuscript's writing.
\bibliography{custom}

\appendix

\section{Extended Results}

\subsection{Ablation}
\label{subsec:extended-ablation}
\textbf{Impact of Network Size.}

\begin{table}[!ht]
\centering
\small
\setlength{\tabcolsep}{3pt}
\resizebox{\columnwidth}{!}{%
\begin{tabular}{ccccc}
\toprule
\shortstack{\textbf{Removal}\\\textbf{(\%)}} & \shortstack{\textbf{F1}\\\textbf{(Macro)}} & \multicolumn{3}{c}{\textbf{Class-wise F1}} \\
\cmidrule(lr){3-5}
& & \textbf{Incorrect} & \shortstack{\textbf{Partially}\\\textbf{Correct}} & \textbf{Correct} \\
\midrule
25 & $0.60\,\pm\,0.01$ & $0.90\,\pm\,0.01$ & $0.32\,\pm\,0.05$ & $0.59\,\pm\,0.02$ \\[3pt]
50 & $0.58\,\pm\,0.02$ & $0.88\,\pm\,0.01$ & $0.30\,\pm\,0.03$ & $0.56\,\pm\,0.02$ \\[3pt]
75 & $0.53\,\pm\,0.01$ & $0.87\,\pm\,0.01$ & $0.25\,\pm\,0.04$ & $0.48\,\pm\,0.01$ \\
\bottomrule
\end{tabular}%
}
\caption{\small{Content-validation performance after randomly removing related videos. Results aggregate $10$ random trials. Removing all related videos corresponds to the \textit{Per-video Pragmatic Inference} baseline in Table~\ref{tab:main-results}.}}
\label{tab:related-video-removal}
\end{table}

Table~\ref{tab:related-video-removal} complements the observational trend in Figure~\ref{fig:predqual_vs_networksize_sensitivity} with a controlled removal test. F1-macro decreases monotonically from $0.60$ to $0.58$ and $0.53$ as we remove $25\%$, $50\%$, and $75\%$ of related videos; removing all related videos yields the per-video score of $0.46$ in Table~\ref{tab:main-results}. Class-wise F1 also decreases at every removal level. Thus, the gain from reflects useful information supplied by the surrounding related videos.

\begin{figure}[t]
  \centering
  \includegraphics[width=\columnwidth]{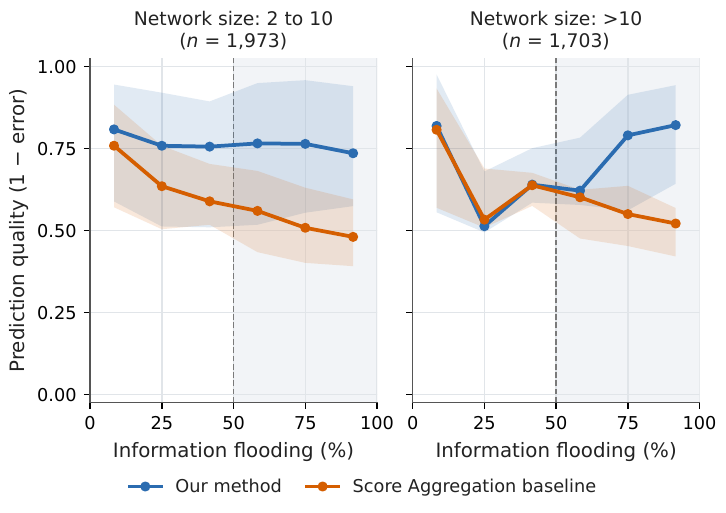}
  \caption{\small{Median prediction quality across information-flooding levels for small and larger networks; shaded regions show the $25\%-75\%$ interquartile range. The dashed line marks the point at which opposing perspectives become the majority.}}
  \label{fig:info-flood-by-size}
\end{figure}

\textbf{Varying Network Composition.}
Figure~\ref{fig:info-flood-by-size} separates the information-flooding experiment in Figure~\ref{fig:info_flood_predqual_vs_networksize} by network size. For networks of size $2$--$10$, our method remains stable across flooding levels. 
% median prediction quality ranges from $0.74$ to $0.81$ . 
In contrast, \textit{Per-video Score Aggregation} declines steadily from $0.76$ to $0.48$. This widening gap shows that our method does not simply conform to the average trust score of the surrounding videos.

For networks larger than $10$, both methods initially lose prediction quality, but they diverge beyond $50\%$ flooding. Specifically, compared to score aggregation baseline, we observe that our method improves at higher information flooding levels. This improvement should not be interpreted as a general benefit of flooding: all targets in this region are labeled \textit{incorrect}, and the larger-network panel contains no \textit{correct} targets. In these cases, additional opposing content can provide contrastive evidence against an incorrect target, which our structured inference can use while score aggregation simply follows the neighborhood average. Subset (or dataset) construction for this experiment is detailed in Appendix~\ref{subsec:impl-ablation}.

\subsection{Analysis of argumentative \& moral framing by credibility}
\label{subsec:argumentative-moral-framing}
% \nish{rewrite}
Table~\ref{tab:discourse-label-definitions} summarizes the discourse dimensions used in Section~\ref{sec:discourse-characterization}. An argumentative strategy is assigned only when a claim has an identifiable inferential role; a claim receives each moral-foundation label that it invokes. Consequently, $28\%$ of claims have no assigned strategy, $31\%$ have no assigned foundation, and claims may contribute to multiple foundation categories.

\begin{figure}[t]
  \centering
  \includegraphics[width=\columnwidth]{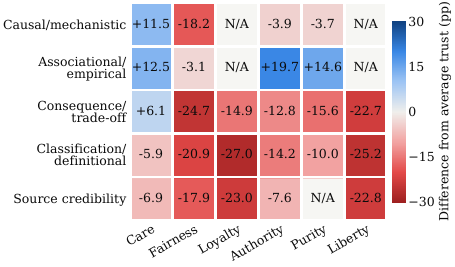}
  \caption{\small{\textbf{Does a strategy receive similar trust regardless of the moral foundation it invokes?} No: the same strategy receives much higher trust when paired with care/harm than with fairness, loyalty, or liberty. Cells show mean claim trust for each strategy--foundation pair relative to the mean over all claims; N/A marks pairs with fewer than $30$ claims. Sign/diagnostic is omitted because most of its pairs fall below this threshold.}}
  \label{fig:strategy-foundation-trust}
\end{figure}

Figure~\ref{fig:strategy-foundation-trust} breaks down claim trust by each combination of argumentative strategy and moral foundation. It supports the finding in Section~\ref{sec:discourse-characterization} that the moral foundation matters more than the strategy: within every strategy, claims invoking \textit{care/harm} are among the most trusted, while those invoking \textit{fairness/cheating}, \textit{loyalty/betrayal}, or \textit{liberty/oppression} are among the least. \textit{Authority/subversion} varies most with the accompanying strategy, ranging from well above average with associational/empirical reasoning to well below average with consequence/trade-off or classification/definitional reasoning, consistent with its use for both citing and challenging institutions (Figure~\ref{fig:authority-by-topic}).

\section{Human Evaluation}
\label{sec:human-evaluation}
We conduct human validation of intermediate model outputs, which were used in dataset construction, claim-prior estimation, and analysis of discourse patterns. At the video level, annotators judge whether the takeaway identified by \texttt{gpt-5.4-mini} captures the video's central message, and whether it aligns with the associated expert-reviewed claim. Together, these two tasks test whether the expert label can be transferred to the video. Annotators are shown the transcript, caption, predicted takeaway, and expert-reviewed claim.

At the claim level, annotators judge the \textit{specificity} and \textit{source invoked}, as predicted by \texttt{Qwen3.5-9B}, which determine the claim prior. They also judge the model predicted argumentative-strategy and moral-foundation assignments used for discourse characterization (Section~\ref{sec:discourse-characterization}). Annotators are shown the claim, takeaway, and all four predicted labels, with access to the transcript, and caption for context. We also provide definitions and examples for each annotation task.

We randomly sample $90$ videos. The two video-level tasks are annotated for every video; for the four claim-level tasks, we randomly select up to three extracted claims per video, yielding $270$ claim items. Each item is independently evaluated by two of the $12$ annotators. The annotators had an average age of approximately $28$, held at least a bachelor's degree in computer science, and were primarily graduate students. Each task asks whether the predicted output is correct or aligned. For the multi-label moral-foundation task, a prediction is correct when at least one assigned foundation is appropriate. Table~\ref{tab:human-evaluation} reports the percentage of positive judgments and raw annotator agreement.

\begin{table}[t]
\centering
\small
\begin{tabular}{@{}lcc@{}}
\toprule
\textbf{Validated output} & \textbf{Result} & \textbf{Agreement} \\
\midrule
Takeaway--video match & $91\%$ & $90\%$ \\
Takeaway--expert alignment & $84\%$ & $80\%$ \\
Source invoked & $87\%$ & $81\%$ \\
Specificity & $93\%$ & $87\%$ \\
Argumentative strategy & $79\%$ & $70\%$ \\
Moral foundations & $82\%$ & $71\%$ \\
\bottomrule
\end{tabular}
\caption{\small{Human validation results and raw inter-annotator agreement. The result column reports the percentage judged correct, aligned, or yes, as applicable.}}
\label{tab:human-evaluation}
\end{table}

At the video level, the extracted takeaway aligns with the expert validation from \texttt{ScienceFeedback} in $84\%$ of cases, with $80\%$ annotator agreement. Moreover, the extracted takeaway itself is judged correct in $91\%$ of videos. Together, these results support the reliability of transferring expert validation to the corresponding influencer content. The \textit{specificity} and \textit{source invoked} annotations similarly support their use in defining claim priors. Argumentative strategies and moral foundations are also validated for most claims, although their lower raw agreement reflects the greater subjectivity of these discourse-level categories.

\section{Implementation Details}
\label{sec:impl_details}
This section reports the settings needed to reproduce our framework and baselines. Prompt templates are provided in Appendix~\ref{sec:all_prompts}.

\subsection{Our Framework}
\label{subsec:impl-framework}

\textbf{LLM settings.}
We use \texttt{gpt-5.4-mini} for \textit{takeaway} and claim extraction, \texttt{Qwen3.5-9B} for the remaining LLM components. For local Qwen inference, we disable extended thinking and use temperature $0.7$, top-$p=0.8$, and the top-$100$ token log-probabilities.
% For a classification with label set $L$, we locate the label token in the generated response and normalize its class log-probabilities over $L$ with a softmax. These probabilities, rather than the generated label alone, determine the edge weights defined in the main paper.

\textbf{Influencer content collection.}
The associated prompts are \ref{prompt:takeaway-extraction}--\ref{prompt:retrieved-video-relevance} in Appendix~\ref{sec:all_prompts}.
We represent each video with its transcript and caption. We transcribe the audio with \texttt{faster-whisper}\footnote{\url{https://pypi.org/project/faster-whisper/}} (\texttt{small.en}). After identifying the \textit{takeaway} (core audience-facing recommendation) for the video, for increased recall using TikTok Research API, we use exactly two short, nonredundant terms--typically an entity and an outcome, for an \texttt{AND} query over TikTok caption metadata. From the obtained topic-specific candidate videos, we extract takeaways and retain a candidate only when its central message addresses the same health issue and can meaningfully reinforce or challenge the target \textit{takeaway}; topical overlap alone is insufficient. Duplicate video IDs and the target video itself are removed.
% This deliberately favors a small, interpretable query over a longer conjunction that would suppress recall.

\textbf{Individual argumentation graphs.} Here, we discuss the details associated with claim extraction, and graph construction.
The associated prompts are \ref{prompt:claim-metadata-extraction}--\ref{prompt:within-video-support} in Appendix~\ref{sec:all_prompts}.

\textit{Claim extraction}: We use the transcript, caption, and previously extracted \textit{takeaway} for this task. The takeaway is supplied only to resolve the video's intended message and prioritize claims relevant to that message; it is not treated as evidence for inventing additional claims.

\textit{Graph construction details}: claims are compared with a target node in batches of five. For estimating edge weight, we normalize the log-probabilities over the classes \{\textit{direct support}, \textit{weak support}, and \textit{none}\}, and compute $P(\mathrm{direct})+0.5P(\mathrm{weak})$. We require a score of at least $0.5$ for claim-to-takeaway edges. This stricter first-level threshold prevents a merely relevant claim from becoming a direct justification for the video's central message. Once a claim is attached, it becomes a target for the remaining unattached claims; recursive claim-to-claim edges use a threshold of $0.1$ so that incomplete but useful intermediate reasoning is not discarded. Expansion ends when no new claim can be attached. Claims that remain unattached receive no within-video support edge.

\textbf{Networked argumentation discourse.}
The associated prompt is \ref{prompt:cross-video-relation} in Appendix~\ref{sec:all_prompts}.
We compare each takeaway with every other takeaway in the same related videos network. For claims, we use \texttt{all-MiniLM-L6-v2}\footnote{\url{https://huggingface.co/sentence-transformers/all-MiniLM-L6-v2}} embeddings and cluster using BERTopic to reduce the number of comparisons with clearly unrelated content. For each claim, we rank cross-video claims in the same semantic cluster by cosine similarity and retain at most the top $20$ candidates, as described in Section~\ref{subsubsec:network-arg-graph}. We retain every non-\textit{none} relation in the network. For the edge score
% including low-confidence edges; its probability-derived score controls its influence during inference instead of imposing another hard threshold.

% Small groups for which clustering is unstable use the same top-$20$ similarity rule. Candidate pairs are then classified in their directed order. 
\textbf{Factual grounding.}
The associated prompts are \ref{prompt:pubmed-checkworthiness}--\ref{prompt:pubmed-evidence-stance} in Appendix~\ref{sec:all_prompts}.
Our retrieval index contains approximately $14$ million English-language PubMed title--abstract records published from 2014 through 2025. We build the index from the 2025 PubMed annual baseline\footnote{\url{https://ftp.ncbi.nlm.nih.gov/pubmed/baseline/}} and use ColBERTv2 for semantic retrieval. This date restriction prioritizes recent biomedical research and guidance while retaining good coverage.
% for longitudinal studies and reviews within the indexed period.

Only claims labeled PubMed-checkworthy enter retrieval. For each such claim, we generate up to five supporting and five opposing queries. We retrieve $20$ documents per query, deduplicate documents by PMID, and retain the top $20$ documents separately, by their retrieval score, for the supporting and opposing pools. An LLM relevance pass removes documents that do not match the claim and the intended pool direction. Abstracts are truncated to $300$ words for this pass. We then classify the remaining articles by stance and retain at most three articles from each pool, preferring strong over weak evidence and excluding neutral or irrelevant articles. A PMID retained from one pool is not duplicated in the other.

\textbf{Trust inference.}
We implement the model with \texttt{pslpython}\footnote{\url{https://pypi.org/project/pslpython/}}. All six rules in Section~\ref{subsubsec:trust-inference} use weight $1$ and linear hinge loss. PubMed nodes are observed at $1$; claim priors are added as observed \textsc{TrustPrior} values through Rules R5--R6; and takeaways are unobserved targets initialized at $0$. We run ADMM~\cite{pmlr-v29-Forouzan13} inference with seed $7$, at most $5{,}000$ iterations. For the reported full model, trust scores in $[0,0.6]$, $(0.6,0.65]$, and $(0.65,1]$ map to \textit{incorrect}, \textit{partially correct}, and \textit{correct}, respectively.

\subsection{Annotated Dataset}
\label{subsec:impl-annotated-dataset}

\textbf{Source alignment.}
We use expert-reviewed health claims from \texttt{ScienceFeedback} as the primary label source. For each expert-claim, we select up to two representative keywords and retrieve candidate videos through the TikTok Research API, restricting the set to English-language videos from creators with at least $1{,}000$ followers. After topical filtering, we extract each video's takeaway and retain the video only when its takeaway is semantically equivalent to the source claim from the expert. This stricter alignment is necessary to  transfer the source verdict to the video's central message.

Due to the class-imbalance issue found in expert-annotated \texttt{ScienceFeedback} claims, especially for claims that receive a \textit{fully correct} verdict, we augment this class using manually curated takeaways from trusted \textbf{\textit{\texttt{MayoClinic}}} videos as gold-standard correct claims. We apply the same retrieval and semantic-equivalence criterion to find TikTok videos expressing those claims; the resulting matches serve as proxy instances of the \textit{correct} class. This procedure contributes $46$ of the final $57$ correct examples.

\textbf{Label mapping and data curation.}
Among the fine-grained verdicts represented in our source claims (from experts), we map \{\textit{inaccurate}, \textit{mostly inaccurate}, \textit{incorrect}, \textit{flawed reasoning}, \textit{unsupported}, and \textit{misleading}\} to \textit{incorrect} class; \{\textit{lacks context}, \textit{imprecise}, \textit{partially correct}, \textit{mostly accurate}, and \textit{mostly correct}\} to \textit{partially correct}; and \{\textit{accurate} and \textit{correct}\} to \textit{correct} class. Overall, the final evaluation set contains $569$ unique labeled videos: $478$ \textit{incorrect}, $34$ \textit{partially correct}, and $57$ \textit{correct}-labeled videos. Table~\ref{tab:dataset-stats} shows the resulting dataset by influencer popularity.

\begin{table}[t!]
\centering
\small
\begin{tabular}{lrr}
\toprule
\textbf{Influencer Popularity} & \textbf{\# Labeled} & \textbf{\# Unlabeled} \\
\midrule
0K--50K & 466 & 442 \\
50K--100K & 43 & 51 \\
$>$100K & 60 & 183 \\
\midrule
\textbf{Total Videos} & \textbf{569} & \textbf{676} \\
% \textbf{Unique Influencers} & \textbf{532} & \textbf{532} \\
\midrule
\textbf{Overall Dataset} & \multicolumn{2}{c}{\textbf{1,245 videos}} \\
\bottomrule
\end{tabular}
\caption{\small{Dataset distribution for labeled (expert-annotated) and unlabeled videos by influencer popularity.}}
\label{tab:dataset-stats}
\end{table}

\subsection{Baselines}
\label{subsec:impl-baselines}

\textbf{Direct prompting.}
The associated prompt is \ref{prompt:direct-content-validation} in Appendix~\ref{sec:all_prompts}.
We provide \texttt{Qwen3.5-9B} only the target transcript and request one of the three task labels with a short rationale. No retrieved evidence, related-video content, extracted takeaway, or gold metadata is included.

\textbf{LOKI.}
The associated prompts are \ref{prompt:loki-claim-extraction}--\ref{prompt:loki-evidence} in Appendix~\ref{sec:all_prompts}.
We adapt LOKI~\cite{li-etal-2025-loki} using their source code\footnote{\url{https://github.com/Libr-AI/OpenFactVerification}}. It extracts a flat set of atomic factual claims from the caption and transcript, verifies each claim independently, and aggregates the results to verify the overall document. To ensure fair comparison across all methods, we retrieve evidence for every query using our indexed PubMed data, rather than relying on \texttt{SERPER-API} as in their source code. Note that we generate two supporting and two opposing retrieval queries \textit{per claim}. Retrieval keeps five hits per query, deduplicates the pooled results by PMID, retains the top five \textit{per claim}, and sends the top three to the evidence-relation classifier. Claim factuality is the fraction of directional evidence labeled \textsc{supports} among evidence labeled \textsc{supports} or \textsc{refutes}; the video score is the mean over claims with directional evidence. Cases with no verified claim remain uncovered rather than receiving an imputed prediction. For this baseline, we conduct a threshold sweep using a grid from $\{0.05, 0.1, ..., 0.95\}$ to identify threshold pairs optimized for obtaining maximum F1-macro score. The reported thresholds are $0.9$ and $0.95$ for three ordered classes. Specifically, $\leq 0.9$ for \textit{incorrect}, $(0.9, 0.95]$ for \textit{partially correct} class, $>0.95$ for \textit{correct} class.

\textbf{Per-video pragmatic inference.}
This baseline uses the same extraction, within-video graph construction, factual grounding, priors, and PSL rules as our full system, but creates one graph containing only the target video. The reported thresholds are $0.55$ and $0.75$ for the three ordered classes, obtained based on a threshold sweep and optimized for maximum F1-macro score.

\textbf{Per-video score aggregation.}
We first run the preceding single-video procedure independently for every video including the target and its related videos in the network, then assign the target the arithmetic mean of their takeaway trust scores. The reported results are with thresholds $0.60$ and $0.70$ (obtained after a threshold sweep, optimized for maximum F1-macro score). Thus, this baseline receives the same videos and scientific-grounding machinery as our method but cannot propagate support, contradiction, or equivalence across videos.

\subsection{Ablation}
\label{subsec:impl-ablation}
\textbf{Varying Network Composition.}
For each labeled target video, we begin with the full related-video graph. Let $A$ contain every other \textit{takeaway}-bearing video in this graph. We partition $A$ into videos whose takeaways are aligned with the target video's perspective and videos whose takeaways contradict it. These judgements are made using an LLM (\texttt{Qwen-3.5-9B}). Each ablation instance retains the target video and a subset $S\subseteq A$. We define its relative size as $(|S|+1)/(|A|+1)$ and its flooding level as $|S_{\mathrm{opp}}|/(|S|+1)$, where the $+1$ accounts for the target video. Thus, $0\%$ flooding contains only \textit{aligned} related videos, while higher values retain an increasingly opposing network. In this manner, we construct different subsets by varying relative size and flooding level, yielding $3{,}676$ subsets for analysis. 
% Because each subset contains only observed videos from the original graph, the ablation changes network composition without synthesizing or relabeling content.

\section{Prompt Specifications}
\label{sec:all_prompts}
We provide the operative prompt instructions below. To keep the appendix compact, we omit illustrative demonstrations and repeated JSON-only output reminders to the model; all prompts preserve the supplied identifiers, and return one decision for every input item. Bracketed text denotes an input field.

\newenvironment{compactprompt}{%
  \begin{quote}\footnotesize\setlength{\parindent}{0pt}\setlength{\parskip}{2pt}%
}{\end{quote}}

\newcounter{prompt}
\renewcommand{\theprompt}{P\arabic{prompt}}
\newcommand{\promptparagraph}[2]{\refstepcounter{prompt}\textbf{\theprompt: #1}\label{#2}}

\subsection{Framework Prompts}
\label{subsec:framework-prompts}

\promptparagraph{Takeaway extraction.}{prompt:takeaway-extraction}
\begin{compactprompt}
\textbf{Input:} [transcript; caption].

\textbf{Instruction:} Extract exactly one central recommendation, warning, suggestion, protocol, or audience-directed implication. Select the message the audience is primarily meant to retain or act upon; if several are equally strong, select the first. Write it as a self-contained, audience-facing statement, resolving pronouns when the input permits. Preserve the original meaning, scope, and certainty. Do not add medical qualifications, factual judgments, or outside knowledge. Return the takeaway, whether it is explicit or implicit, a short selection rationale, and the exact grounding span. If the input contains no coherent takeaway, return \texttt{NO\_TAKEAWAY\_FOUND}.
\end{compactprompt}

\promptparagraph{Search-keyword selection.}{prompt:search-keyword-selection}
\begin{compactprompt}
\textbf{Input:} [takeaway; optional caption/transcript context].

\textbf{Instruction:} Select exactly two simple, atomic terms for an AND search over social-media captions. Prefer common words likely to occur in captions, usually one for the main intervention/entity and one for the condition, outcome, or core issue. Avoid redundant variants, unnecessary modifiers, and overly specific phrases. Balance recall with precision. Optionally return one auxiliary term for later filtering, but do not place it in the search conjunction.
\end{compactprompt}

\promptparagraph{Retrieved-video relevance.}{prompt:retrieved-video-relevance}
\begin{compactprompt}
\textbf{Input:} [target takeaway; candidate takeaway].

\textbf{Instruction:} Decide whether the candidate addresses the same underlying health issue as the target and expresses a message that is equivalent to, supports, or opposes the target. Recommendations may differ, but the overlapping factual issue must be identifiable. Shared entities or topic words alone are insufficient. Return \texttt{relevant} with the applicable relation, or \texttt{not\_relevant}, with one-sentence justification.
\end{compactprompt}

\promptparagraph{Claim and metadata extraction.}{prompt:claim-metadata-extraction}
\begin{compactprompt}
\textbf{Input:} [transcript; caption; takeaway].

\textbf{Instruction:} Extract every atomic, testable proposition stated or strongly implied in the transcript or caption that helps characterize the video's message. Each claim must be self-contained, preserve concrete qualifiers (population, dose, duration, comparison, source, and outcome), and include an exact grounding span. Do not invent a claim from the takeaway. Label each claim as explicit or implicit; specificity as \texttt{high} only when concrete bounded details materially narrow the proposition, otherwise \texttt{low}; and the primary source invoked as \texttt{institutional/scientific evidence}, \texttt{experiential evidence}, \texttt{social testimony}, or \texttt{unsupported assertion}. These source labels describe presentation, not truth. Return an empty list when no claim is present.
\end{compactprompt}

\promptparagraph{Within-video support.}{prompt:within-video-support}
\begin{compactprompt}
\textbf{Input:} [target claim; candidate statements from the same video].

\textbf{Instruction:} Classify the directed relation from each statement to the target claim as \texttt{direct support}, \texttt{weak support}, or \texttt{none}. Direct support clearly entails, restates, justifies, or gives evidence for the target. Weak support makes the target more credible but requires an additional inference. Use none for topical overlap, missing assumptions, or a highly indirect connection. Do not infer unstated facts. Return a one- or two-sentence rationale for each statement.
\end{compactprompt}

\promptparagraph{Cross-video relation.}{prompt:cross-video-relation}
\begin{compactprompt}
\textbf{Input:} [source claim; candidate claims from other videos].

\textbf{Instruction:} Classify the directed relation from the source to each candidate as \texttt{equivalent}, \texttt{contradicts}, \texttt{supports}, or \texttt{none}, in that decision order. Equivalent requires the same core proposition, polarity, entities, and material scope. Contradiction requires conflicting assertions about the same issue. Support requires the source to provide a direct reason for the candidate and is directional. Use none for compatible statements about different aspects, reverse-direction support, vague relatedness, or uncertainty. Use only the supplied text and provide one sentence of rationale.
\end{compactprompt}

\promptparagraph{PubMed checkworthiness.}{prompt:pubmed-checkworthiness}
\begin{compactprompt}
\textbf{Input:} [transcript; extracted claims].

\textbf{Instruction:} Label each claim \texttt{PubMed-checkworthy} when it makes a biomedical or health-related factual assertion that is generalizable beyond the immediate anecdote and can in principle be evaluated using PubMed literature. Label personal, rhetorical, non-biomedical, unresolvably vague, or context-specific statements \texttt{not PubMed-checkworthy}. Judge verifiability rather than truth; when genuinely uncertain, prefer checkworthy. Give one brief explanation per claim.
\end{compactprompt}

\promptparagraph{Bidirectional query expansion.}{prompt:bidirectional-query-expansion}
\begin{compactprompt}
\textbf{Input:} [transcript; PubMed-checkworthy claim].

\textbf{Instruction:} Generate up to five supporting and five opposing semantic-search queries. Each query must be a concise, standalone natural-language phrase or question suitable for embedding retrieval, without Boolean syntax or field tags. Cover distinct, claim-compatible angles such as outcome, mechanism, population, route, study design, safety, adverse events, or null effects. Opposing queries should seek contrary, null, or adverse evidence. Preserve the claim's population, intervention, and outcome; do not introduce a new effect or broaden its scope. Return a one-sentence justification for each query.
\end{compactprompt}

\promptparagraph{PubMed relevance filtering.}{prompt:pubmed-relevance-filtering}
\begin{compactprompt}
\textbf{Input:} [takeaway; claim; intended supporting/opposing pool; article titles and abstracts].

\textbf{Instruction:} For each article, first determine whether it addresses a matching exposure/intervention, outcome, population, or comparison; then determine whether its reported findings could serve the intended pool direction. Return \texttt{relevant} only when the article is useful for evaluating this claim in that pool. Return \texttt{not\_relevant} for off-topic work, mismatched outcomes, insufficient evidence, or an article that belongs to the opposite pool. Do not treat keyword overlap or retrieval rank as evidence, and do not infer unstated results. When uncertain, choose not relevant.
\end{compactprompt}

\promptparagraph{PubMed evidence stance.}{prompt:pubmed-evidence-stance}
\begin{compactprompt}
\textbf{Input:} [claim; article titles and abstracts].

\textbf{Instruction:} Classify each article as \texttt{SS} (direct, strong support), \texttt{WS} (limited, indirect, qualified, or context-specific support), \texttt{WO} (limited, indirect, qualified, or context-specific opposition), \texttt{SO} (direct, strong opposition), \texttt{N} (related but nondirectional), or \texttt{I} (irrelevant). Evaluate the exact claim and distinguish population, dose, route, timing, comparison, and outcome. Use strong labels only for directly aligned evidence. For a negative claim, a null result supports it only when the article evaluates the same effect. Use only the title and abstract and give one concise rationale.
\end{compactprompt}

\subsection{Baseline Prompts}
\label{subsec:baseline-prompts}

\promptparagraph{Direct content validation.}{prompt:direct-content-validation}
\begin{compactprompt}
\textbf{Input:} [target transcript].

\textbf{Instruction:} Assess the transcript as a whole using general medical and scientific knowledge. Return \texttt{fully\_correct} when all material claims and recommendations are accurate without a consequential missing qualification; \texttt{partially\_correct} when accurate material content co-occurs with an important error, overstatement, unsupported generalization, or missing qualification; and \texttt{incorrect} when the central claim or recommendation is false, unsupported, or substantially misleading and any accurate details are incidental. Preserve distinctions in population, timing, dose, route, comparison, certainty, and outcome. Return the label and a concise rationale.
\end{compactprompt}

\textbf{P12--P15: LOKI adaptation.}
\begin{compactprompt}
\refstepcounter{prompt}\label{prompt:loki-claim-extraction}
\textbf{Claim extraction input:} [caption; transcript]. Extract concise, atomic, independently checkable health claims. Resolve references using only the input; preserve material qualifiers; split independently assessable conjunctions; remove semantic duplicates; and exclude recommendations, opinions, rhetorical questions, and purely anecdotal experiences unless they contain a concrete factual assertion.

\refstepcounter{prompt}\label{prompt:loki-checkworthiness}
\textbf{Checkworthiness input:} [claims]. Mark each claim \texttt{CHECKWORTHY} when PubMed evidence could objectively support or refute its empirical medical, biological, epidemiological, treatment, prevention, diagnostic, or risk assertion. Otherwise mark \texttt{NOT\_CHECKWORTHY}. Judge investigability, not truth.

\refstepcounter{prompt}\label{prompt:loki-query}
\textbf{Query input:} [checkworthy claims]. For each claim, generate exactly two supporting-direction and two opposing-direction sentence-like PubMed queries. Use established terminology and retain material population, intervention/exposure, and outcome constraints. Do not repeat the original claim, use Boolean fragments, assume a truth label, omit a claim, or invent one.

\refstepcounter{prompt}\label{prompt:loki-evidence}
\textbf{Evidence input:} [claims; retrieved article titles/abstracts]. For every claim--article pair, return \texttt{SUPPORTS} when the evidence directly backs the claim at a compatible scope, \texttt{REFUTES} when it directly contradicts a material part or reports a compatible null/opposite result, and \texttt{IRRELEVANT} when it is indirect, mismatched, or insufficient. Use only the supplied evidence and give a brief rationale.
\end{compactprompt}

\end{document}